\documentclass[letterpaper, 10 pt, conference]{ieeeconf}  

\IEEEoverridecommandlockouts                              

\usepackage{graphicx} 
\usepackage[utf8]{inputenc}
\usepackage{amsmath,amsfonts,booktabs,cite} 
\usepackage{gensymb} 
\usepackage{siunitx,textcomp} 
\usepackage[hidelinks]{hyperref} 
\usepackage{cleveref} 
\usepackage{subcaption}
\usepackage{tabularx}
\usepackage{pbox}
\let\labelindent\relax
\usepackage[inline]{enumitem} 
\usepackage[nolist]{acronym} 
\usepackage{multirow}
\usepackage[ruled,vlined]{algorithm2e}

\usepackage[font=small]{caption}

\def\secref#1{Sec.~\ref{#1}}
\def\figref#1{Fig.~\ref{#1}}
\def\tabref#1{Tab.~\ref{#1}}
\def\eqref#1{Eq.~(\ref{#1})}
\def\algref#1{Alg.~\ref{#1}}

\newcommand\etal{~\emph{et al. }}
\newlength{\twosubht}
\newsavebox{\twosubbox}

\crefname{algocf}{alg.}{algs.}
\Crefname{algocf}{Algorithm}{Algorithms}

\title{\LARGE \bf
Viewpoint Planning for Fruit Size and Position Estimation
}

\author{Tobias Zaenker \and  Claus Smitt \and  Chris McCool \and Maren Bennewitz
\thanks{This work was funded by the Deutsche Forschungsgemeinschaft (DFG, German Research Foundation) under Germany’s Excellence Strategy – EXC 2070 – 390732324. All authors are with the University of Bonn, Germany.}}

\begin{document}

\maketitle
\thispagestyle{empty} 
\pagestyle{empty}

\begin{abstract} 

Modern agricultural applications require knowledge about the position and size of fruits on plants.
However, occlusions from leaves typically make obtaining this information difficult.
We present a novel viewpoint planning approach that builds up an octree of plants with labeled regions of interest~(ROIs), i.e., fruits.
Our method uses this octree to sample viewpoint candidates that increase the information around the fruit regions and evaluates them using a heuristic utility function that takes into account the expected information gain.
Our system automatically switches between ROI targeted sampling and exploration sampling, which considers general frontier voxels, depending on the estimated utility.
When the plants have been sufficiently covered with the \mbox{RGB-D} sensor, our system clusters the ROI voxels and estimates the position and size of the detected fruits.
We evaluated our approach in simulated scenarios and compared the resulting fruit estimations with the ground truth.
The results demonstrate that our combined approach outperforms a sampling method that does \textit{not} explicitly consider the ROIs to generate viewpoints in terms of the number of discovered ROI cells.
Furthermore, we show the real-world applicability by testing our framework on a robotic arm equipped with an \mbox{RGB-D} camera installed on an automated pipe-rail trolley in a capsicum glasshouse.
\end{abstract} 

\section{Introduction}
\label{sec:intro}

Advanced automated agricultural applications such as fruit picking or targeted crop spraying require spatial information about plants.
With a fixed sensor setup, generating a 3D~model can be difficult, as parts of the plants are typically occluded.
Especially plants with a large number of leaves are challenging.
Therefore, having a mobile sensor placed on a robotic arm to move it around and avoid occlusions is a promising solution.
However, in order to know where to place the sensor, viewpoints have to be planned first.

Conventional viewpoint planning approaches usually aim for building complete 3D models of the environment \cite{pito1996sensor, palazzolo2018effective, monica2018contour}.
This can be difficult to achieve for plants due to their complex structure with leaves that occlude fruits.
However, often spatial knowledge about certain regions of interest~(ROIs) is enough.
For example, for fruit picking, accurate information about the location and size of fruits is required, but the exact shape of every leaf is not needed.

\begin{figure}
	\sbox\twosubbox{
		\resizebox{\dimexpr.95\linewidth}{!}{
			\includegraphics[height=3cm]{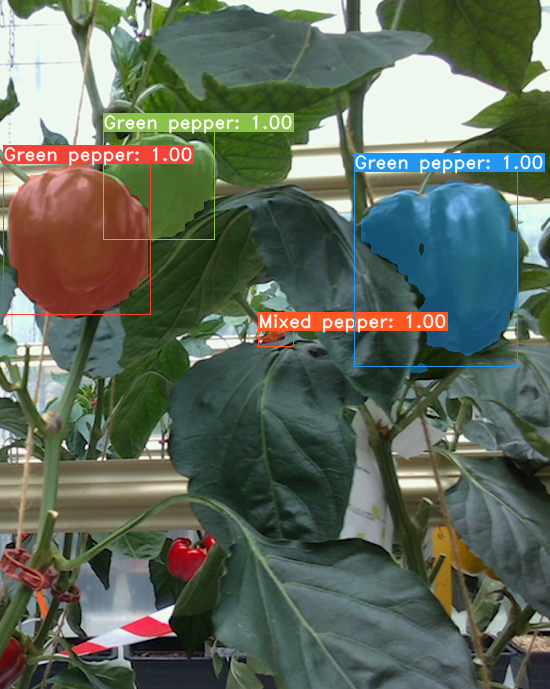}
			\includegraphics[height=3cm]{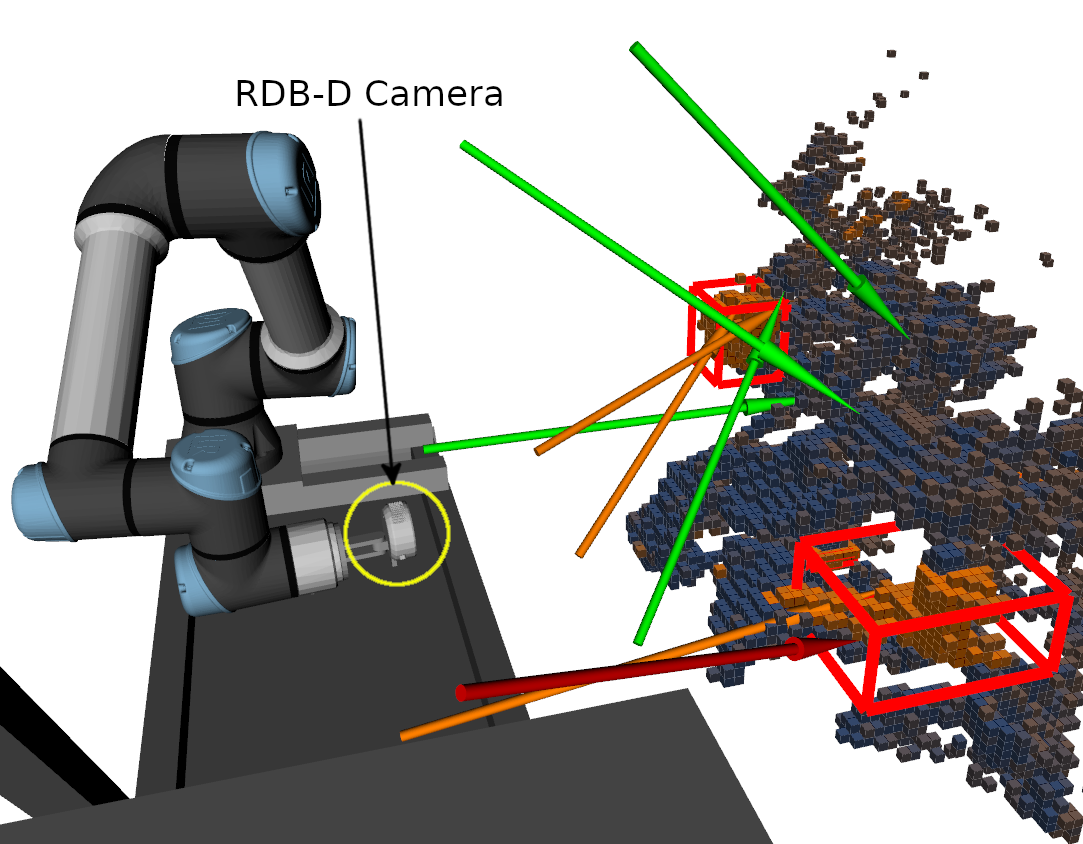}
		}
	}
	\setlength{\twosubht}{\ht\twosubbox}
	\centering
	\subcaptionbox{Detected fruits\label{fig:detection_example}}{
		\includegraphics[height=\twosubht]{images/detection_example}
	}\quad
	\subcaptionbox{Viewpoint selection\label{fig:planning_example}}{
		\includegraphics[height=\twosubht]{images/planning_example_2}
	}
	
	\caption{Illustration of our approach. \textit{Left}: Fruits are recognized in the image but partially occluded, which makes size estimation difficult. \textit{Right}: Detected fruits~(red boxes) marked in the 3D representation are used to sample viewpoints~(indicated by orange arrows) that increase the information around the fruit regions. Additionally, we also sample viewpoints for general frontier voxels independently of the detected fruits to further explore the plant and detect new fruits~(green arrows). The red arrow corresponds to the viewpoint with the highest utility, which is selected as
    the next viewpoint and sent to the robot.
    }
	\label{fig:coverfig}
\end{figure}

In this paper, we present a novel viewpoint planning approach that detects objects, e.g., fruits, online during planning and marks ROIs in a 3D representation.
The detected ROIs are then used to sample viewpoints that increase the information around them.
We combine this method with an exploration approach that samples viewpoints for frontier voxels independently of the detected ROIs to also detect new ROIs in so far unexplored regions.
\figref{fig:coverfig} illustrates our proposed approach.
First, fruits are detected in the image, and the corresponding ROIs are marked in the 3D representation.
Note that some fruits are only partially visible due to occlusions.
Subsequently, our system samples viewpoints that provide a view on the fruits from alternative directions in order to increase the spatial knowledge about them.

We implemented a customized octree structure on top of the OctoMap framework~\cite{hornung13auro} that stores information about occupancy and ROI probabilities.
Our system automatically switches between ROI targeted and exploration sampling depending on the stage of plant coverage.
For each viewpoint, the planner computes a utility value based on the expected information gain to determine the best view of the sampled candidates.
The source code of our system is available on GitHub\footnote{\url{https://github.com/Eruvae/roi_viewpoint_planner}}.
Our main contributions are the following: 
\begin{itemize}
	\item A novel next best view approach for viewpoint planning that detects and uses regions of interest,
	\item A method to estimate the location and size of fruits,
	\item An evaluation of the planner in simulated scenarios,
comparing our combined view pose sampling approach to a method that samples at frontiers to unknown space without considering ROIs, and using two different utility functions, 
	\item Implementation of the approach on a real-world robot platform and demonstration of its use in a commercial glasshouse environment.
	
\end{itemize} 

\section{Related Work}
\label{sec:related}

Viewpoint planning approaches can be divided into coverage path planners (CPP), which compute a complete viewpoint path covering the desired area of a known map, and local next best view (NBV) planners, which are used for unknown environments.
Most CPP approaches hereby assume a static environment.
For example, \mbox{O{\ss}wald\etal\cite{oswald_efficient_2017}} generate viewpoints by casting rays from known object voxels towards free space and evaluate them based on the number of visible object voxels for randomly sampled directions.
For all view poses that exceed a given utility threshold, a robot configuration to obtain that pose with minimal cost is computed.
Finally, the authors apply a traveling salesman problem solver to compute the smallest tour of viewing poses that cover all observable object voxels.
Jing\etal\cite{jing_coverage_2019} generate viewpoints based on the maximum sensor range and compute viewing directions from the surface normals of all target voxels within a certain range.
The authors propose to sample a random set of points and connect nearby points to a graph with a local planner.
Starting with the current robot pose, the neighbors with the highest ratio of expected information gain~(IG) and move cost are added to the solution path until the desired coverage is reached.
CPP has a wide variety of use-cases, from planning the path for cleaning robots~\cite{de1997complete} to covering an agricultural field with machines for crop farming~\cite{oksanen2009coverage}.
However, CPP approaches require a given representation of the environment, which is not applicable for our agricultural use-case, as the environment can change rapidly with the growth of the plants.
Therefore, our approach does not assume a map to be given but builds it during operation.

NBV approaches either use only current sensor information or build a map of the environment while traversing it and use it to decide on the next view.
An example for the former approach was presented by Lehnert\etal\cite{lehnert_3d_2019} who use an array of cameras and determine the size of a target in each frame.
The authors propose to compute a gradient to determine the direction for which the visible area of the target is increased.
Wang\etal\cite{wang_autonomous_2019} use both current sensor information and a built map for planning and propose to combine an entropy-based hand-crafted metric, which is computed by tracing rays through the generated map, with a metric learned by a CNN that only takes the current depth image as input.
The two metrics are combined to evaluate candidate poses generated in the vicinity of the current camera position.
While such approaches are useful to avoid local occlusions, coverage of larger environments is not straightforward.

In the approach proposed by Monica\etal\cite{monica_humanoid_2019}, the task of the robot is to explore the environment around a single object of interest with a known pose while performing a 3D~shape reconstruction of the initially unknown environment.
The authors also apply an exploration behavior for unknown parts of the environment, mainly to find new paths that may enable observations of the object of interest.
Viewpoints are sampled either around the target or at the frontier to unknown space and the viewpoint with the highest IG is then chosen as the next best view.
Similarly, \mbox{Palazzolo\etal\cite{palazzolo2018effective}} sample viewpoints on the hull of the currently known map and select the best point based on the estimated utility taking into account the expected IG.
\mbox{Bircher\etal \cite{bircher2016receding}} propose to use a rapidly exploring random tree and estimate the exploration potential based on the unmapped volume that can be explored at the nodes along the branches of the tree.
Only the first segment of the best branch is then executed and the IG is reevaluated based on the newly gathered data.

Monica\etal\cite{monica2018contour} presented an NBV method that samples viewpoints from general frontiers to unknown space.
We use a similar technique to sample viewpoints for unexplored regions independently of the detected ROIs when those have been sufficiently explored, and provide a comparison to the sampling based on~\cite{monica2018contour} for our application.

Similar to the approach of Sukkar\etal\cite{sukkar_multi-robot_2019} who detect apples as ROIs through color thresholding, we also rely on an automatic detection of the ROIs for viewpoint planning.
Sukkar\etal propose to evaluate viewpoints based on a weighted sum of exploration information, which is calculated from the number of visible voxels that have not been previously explored, and ROI information, which evaluates the visibility of ROIs from the selected viewpoints.
This evaluation metric is then used to plan a sequence of viewpoints for multiple robot arms by utilizing a decentralized Monte Carlo tree search algorithm.
This approach is the most similar one to ours, however, instead of using the ROIs for evaluating view poses in an integrated path planner, we sample view candidates from the detected regions.
While this may result in a less accurately estimated IG, as only a single point instead of a complete path is evaluated, the complexity is lower, and it can easily be used as high-level goal planner for any existing motion planner.

All NBV approaches heavily rely on metrics to evaluate the candidate views.
In our evaluation, we utilize and combine different metrics proposed and evaluated by Delmerico\etal\cite{delmerico18ijrr} to determine the best next view.

\section{System Overview}
\label{sec:approach}

Our viewpoint planning approach aims at finding viewpoints that improve the knowledge about specified regions of interest (ROIs), i.e., the fruits of plants for our application.
In order to do that, the ROIs have to be detected in the current field of view and marked in the planning map.
The marked ROIs are then taken into account for sampling and evaluation of new viewpoint candidates and are used to estimate the size and location of fruits in the final map.

\begin{figure*}[t] 	\centering 	\includegraphics[width=\linewidth]{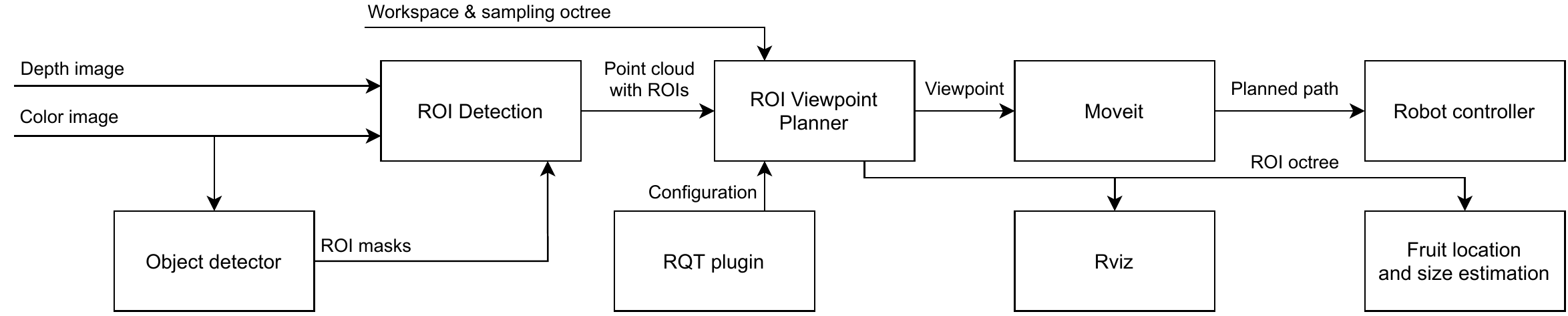} 	\caption{Overview of our system.
See text for a detailed description.} 
	\label{fig:planner_overview}
\end{figure*} 

\Cref{fig:planner_overview} shows an overview of our framework.
The depth and color image from an RGB-D sensor are used as input for the planner.
The color image is forwarded through an object detection network, which outputs masks of the ROIs.
The detection module uses these masks in combination with the depth image to generate a point cloud divided into ROI and non-ROI regions (\secref{sec:approach_roidet}).
The generated point cloud is forwarded to the planner module, which uses the incoming information to build up a 3D map of the environment in the form of an octree, which stores both occupancy and ROI information (\secref{sec:approach_map}).
From the generated map, targets and viewpoints are sampled and evaluated (\Cref{sec:approach_sampling}).
The ROI octree is also used to evaluate the viewpoint selection and estimate fruit locations and sizes.
Furthermore, we provide a sampling and a workspace octree as input to specify the regions where valid targets and viewpoints can be sampled respectively.
An RQT plugin provides a graphical user interface to configure planning parameters as well as to save and load the generated octrees.

Using the MoveIt framework \cite{chitta2012moveit}, our system plans a path for the robot to the best found viewpoint.
The planned path is then executed by the robot controller.

\section{Preliminaries}

\subsection{ROI Detection}
\label{sec:approach_roidet}

To plan viewpoints that improve information around fruits as well as to estimate their size and position, the fruits have to be detected.
Since the plants in our simulation experiments only had red peppers, we employed a simple color detection.
The detection module first downsamples the colored point cloud to the octree resolution using a voxel grid filter.
Then, it performs a color comparison in the HSI color space.
Points with a hue between -30° and 50° as well as saturation and intensity values of at least 0.12 are considered as fruit points.

For the real-world experiments, peppers of multiple colors were present, including green peppers.
Since green peppers cannot be detected using color filters due to the green leaves, we used a different approach.
The fruit detection is handled by the neural network Yolact~\cite{bolya2019yolact}, which is an instance segmentation network that predicts bounding boxes and masks of objects in the color images.
We trained the network using a dataset recorded at the University of Bonn~\cite{halstead2020fruit}.

Next, the detections have to be transferred to 3D space in order to build a map.
Therefore, we use the point cloud generated from the color-aligned depth images of the \mbox{RGB-D} camera.
The point cloud is downsampled to the resolution of the octree using a voxel grid filter.
Points within the masks of a detected fruit are marked as ROI.

\subsection{Octree for Viewpoint Planning}
\label{sec:approach_map}

In order to build a 3D representation that is able to mark and update ROIs, we use a custom octree, where each node stores two values: an occupancy and a ROI probability, both stored as log-odds.
The octree is updated using a point cloud with marked ROIs and the sensor origin of the point cloud.
The implementation is built on top of the popular OctoMap framework \cite{hornung13auro}.

First, the occupancy is updated by casting rays from the sensor origin to each of the points in the point cloud.
All nodes that are traversed on the way to the points are added to a set of free nodes, while the point itself is added to the set of occupied nodes.
After all points are processed, nodes that are in both the free and the occupied set are removed from the free set.
Then, the occupancy log-odds of the free nodes are reduced and the log-odds of the occupied nodes are increased.
We also use an upper and lower bound for the log-odds to limit the confidence for a node.
Nodes with positive log-odds are considered as occupied, nodes with negative log-odds as free.

The ROI probability is updated in a similar fashion.
Here, only nodes directly corresponding to points in the point cloud are used.
For all points marked as ROI, the corresponding node is added to the set of ROI nodes.
For all other points, the nodes are added to the set of non-ROI nodes, if they are not already part of the ROI node set.
Then, the ROI log-odds are increased for the ROI node set and decreased for the non-ROI node set.
Once the ROI log-odds surpass a set threshold, the nodes are considered as ROI nodes.

\section{Viewpoint Planning}
\label{sec:approach_sampling}

We now describe our novel approach of viewpoint planning that takes into account ROIs for sampling viewpoint candidates.
In particular, we developed a method that uses a combination of two sampling methods: a ROI targeted method improving the information around already discovered ROIs, and an exploration method targeting general frontier voxels to find new ROIs in so far unexplored regions.
We hereby use two additional octrees to sample valid view poses, a \textit{workspace tree} and a \textit{sampling tree}.
Both our sampling methods sample their targets from the region specified by the sampling tree, while sampled viewpoints are only considered if they lie within the workspace tree: \begin{description} 
\item[Workspace Octree]
To initially generate the workspace octree, we loop through a discrete set of joint configurations and store possible view poses.
For the used robot, additional constraints can be taken into account, e.g., for the glasshouse experiments, our robotic arm should only move within a specified corridor to not damage the plants, so we consider the corresponding limits.
\item[Sampling Octree]
For the sampling octree, we use two different approaches in our experiments.
For the simulation, we simply use the workspace octree, as the plants are within the workspace of the arm.
For the real-world experiments, the target plants were on one side of the corridor, outside of the workspace.
To adjust for that, we inflated the workspace by $60\,cm$, and cut off the parts outside of the plant row.
\end{description}

\subsection{ROI Targeted Sampling}
\label{sec:sampling_roi}

For the ROI targeted sampling, our system determines frontiers in the vicinity of detected ROIs, i.e., the \mbox{6-neighborhood} of all ROI nodes of the planning octree within the sampling octree is checked for free nodes.
All free nodes that additionally have an unknown neighbor are considered as ROI frontiers, and therefore potential targets.

After all targets are generated, possible viewpoints are sampled in all directions at random within a specified sensor range.
If these viewpoints lie within the workspace octree, they are further processed, i.e., the orientation is determined by rotating the camera pose so that the viewing direction aligns with the vector from the viewpoint to the target.
Furthermore, a ray is cast between the viewpoint and the target point.
If the ray passes an occupied node, the viewpoint is discarded, as the target is occluded.

\subsection{Exploration Sampling}
\label{sec:sampling_explo}

To be able to find new ROIs after all current ROIs have been sufficiently explored, we implemented a second sampling method that considers general frontier voxels.
Similar to \mbox{Monica\etal\cite{monica2018contour}}, our approach looks for frontiers at the border of occupied and unexplored space, instead of just the frontiers of already detected ROIs.
To find such voxels, we check all free nodes of the planning octree that lie within the sampling octree.
If the 6-neighborhood contains both an occupied and an unknown node, it is considered as a potential target.
After all targets are collected, potential viewpoints are sampled and their directions are determined in the same way as for the ROI targeted sampling.

\subsection{Viewpoint Evaluation}
\label{sec:approach_evaluation}

For the sampled view poses, we need to estimate the information gain~(IG).
To do so, we cast rays from the view pose within a specified field of view of the sensor.
For each ray, we estimate the IG based on metrics presented by Delmerico\etal\cite{delmerico18ijrr}.
First, we compute the Unobserved Voxel IG, where all voxels that have not been encountered so far contribute to the information gain.
For each ray $r$, the number of unknown voxels $N_{u, r}$ is counted starting from the origin, until either an occupied voxel is encountered or the end of the specified sensor range is reached.
The IG for this ray is then computed as $N_{u, r}$ divided by the total number of nodes~$N_r$ on the ray.
For the total IG of a view pose, the average IG of all rays is computed: 
\begin{align}
IG_U = \frac{1}{\vert R\vert}\sum_{r\in R}\frac{N_{u, r}}{N_r}
\end{align}

The second metric is similar to the Proximity Count in~\cite{delmerico18ijrr}, which was one of the metrics that performed best in their comparison.
In the original formulation, unobserved voxels are weighted higher if they are close to an already observed surface.
For our approach, we slightly modified this metric.
Instead of increasing the weight for voxels close to all surfaces, we only do so for voxels close to observed ROIs.
Each unknown voxel is given an initial weight of 0.5, and if it is within a specified distance~$max\_dist$ from a known ROI, the weight~$w$ is computed as follows: 
\begin{align}
	w = 0.5 + 0.5 \cdot\frac{\mathit{max\_dist} - \mathit{dist}}{\mathit{max\_dist}}
\end{align}
where~$\mathit{dist}$ is its distance to the ROI. Known voxels receive a weight of 0.
Considering the sum of the weights of all voxels on a ray $W_r$, the computation of the information gain changes to 
\begin{align}
	IG_P = \frac{1}{\vert R\vert}\sum_{r\in R}\frac{W_r}{N_r}
\end{align}

In addition to the IG, we compute the cost $C$ for reaching the viewpoint.
Since computing the joint trajectory to reach the viewpoint is too time-consuming to be done for every viewpoint, we use the Euclidean distance of the camera to the point as an approximation.
Finally, the utility of a view pose is computed as the weighted sum of IG and the cost scaled by a factor~$\alpha$: 
\begin{align}
U = IG - \alpha \cdot C
\end{align}

\begin{algorithm}[t]
	\SetAlgoLined
	\While{True}{
		roiVps = sampleRoiVps(pose, nVps, utilType)\;
		explVps = sampleExplVps(pose, nVps, utilType)\;
		makeHeap(roiVps, explVps); // sorted by utility\\
		\eIf{max(roiVps) $>$ utilityThreshold}{
			chosenVps = roiVps\;
		}{
			chosenVps = explVps\;
		}
		\While{max(chosenVps) $>$ utilityThreshold}
		{
			vp = extractMax(chosenVps)\;
			\If{moveToPose(vp)}
			{
				break\;
			}
		}
	}
	\caption{Viewpoint planning}
	\label{algo:vpp}
\end{algorithm}

\subsection{Viewpoint Selection}

In our proposed approach, we combine both methods, ROI targeted sampling and exploration sampling, for generating candidates.
\algref{algo:vpp} describes the basic structure of our viewpoint planning approach.
In the planner loop, we sample viewpoints using ROI and exploration sampling, described in \secref{sec:sampling_roi} and \secref{sec:sampling_explo} respectively.
The planner then generates a max heap from the sampled viewpoints, sorted by a utility value, introduced in \secref{sec:approach_evaluation}.
If the maximum utility of the ROI viewpoints is above an empirically found threshold, they are used.
Otherwise, the planner switches to the exploration viewpoints and checks whether their maximum utility is above the threshold.
The planner then tries to plan a path to the viewpoint with the maximum utility.
If no collision-free path can be found, the next best viewpoint is used, until either the planner is successful, or no viewpoint above the utility threshold is left in the heap.
In the latter case, new view poses are sampled.

\section{Experiments}
\label{sec:exp}

We evaluated our planning approach with an RGB-D camera placed on a robotic arm.
We used a UR5e from Universal Robots, both for the simulated scenarios and for the real-world experiments.
The arm has six degrees of freedom and a reach of $85\,cm$.
To compute the workspace of the arm, the first 5 joints were sampled at a resolution of $10^{\circ}$.
The 6th joint was ignored, as it only rotates the camera and therefore does not change the viewpoint.
The collision-free poses were marked in the workspace octree with a resolution of $2\,cm$.
We set the planning tree resolution to $1\,cm$, which provided the best compromise of planning time and map resolution.
In each planning iteration, 100 ROI and 100 exploration viewpoints were sampled.

To identify individual fruits, we first cluster the identified ROI nodes and then estimate their position and volume.
We sequentially process the identified ROI nodes and inspect their complete 26-neighborhood of voxels.
Any found ROI node is added to the current cluster and pushed to a list of voxels to be processed.
The cluster is expanded until no more neighbors are found.
If any ROI nodes are left, a new cluster is started.
After all clusters are computed, they can be used to estimate the fruit position and size.
In our experiments, we calculate the average coordinate of all nodes in a cluster as fruit position, and the volume of the 3D bounding box as an estimate for the size of the fruits.

\subsection{Simulated Scenarios}

\begin{figure}
	\sbox\twosubbox{
		\resizebox{\dimexpr.98\linewidth}{!}{
			\includegraphics[height=1.8cm]{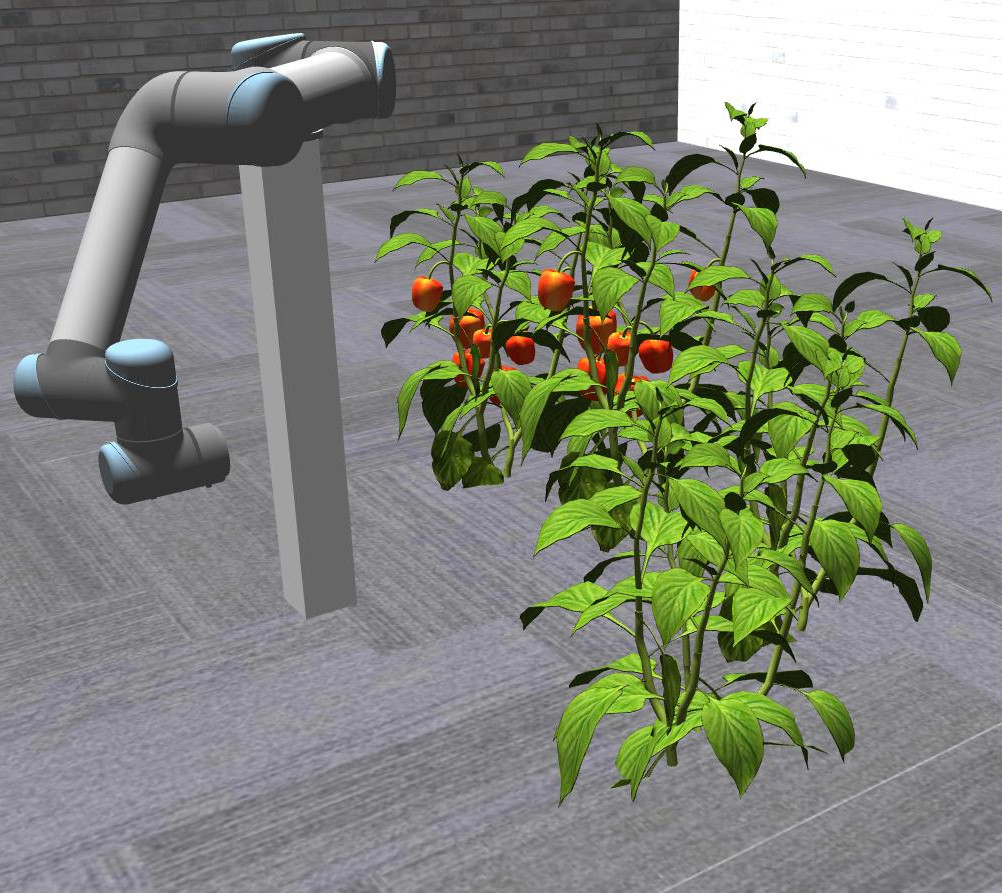}
			\includegraphics[height=1.8cm]{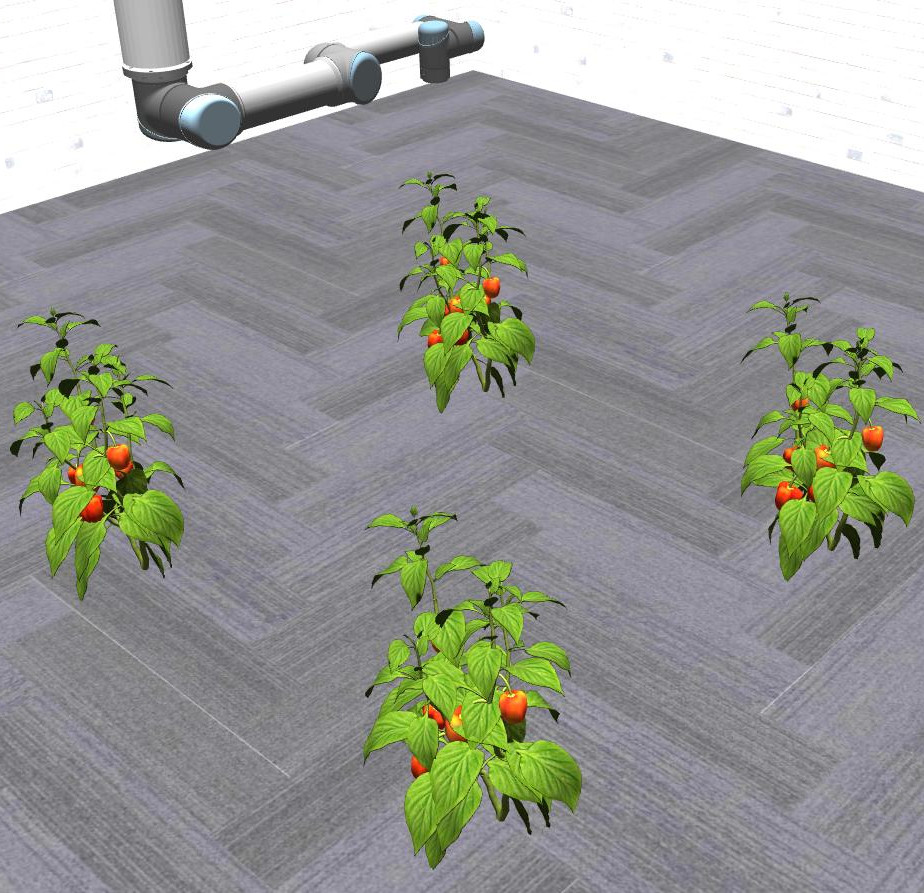}
		}
	}
	\setlength{\twosubht}{\ht\twosubbox}
	\centering
	\subcaptionbox{\textit{Scenario 1}\label{fig:simulatedenv1}}{
		\includegraphics[height=\twosubht]{images/simulated_env_1_cropped}
	}\quad
	\subcaptionbox{\textit{Scenario 2}\label{fig:simulatedenv2}}{
		\includegraphics[height=\twosubht]{images/simulated_env_2_cropped}
	}
	\caption{\textit{Scenario 1:} Simulated environment with four plants, two of which have seven fruits each and the other two do not have any fruits. The arm is placed on a static pole.
	\textit{Scenario 2}: More complex simulated environment containing four plants with seven fruits each. The arm is hanging from
	the ceiling and can move within a $2\times2\,m$ square and extend up to $1.2\,m$ down.
	}
\end{figure}

Two environments with different workspaces were designed for the simulated experiments.
In the first scenario, the arm is placed on top of a static $85\,cm$ high pole~(see \figref{fig:simulatedenv1}).
This allows the arm to exploit most of its workspace, except for the part blocked by the pole.
However, the movement possibilities are limited, as the arm cannot move itself.
To be able to explore a larger workspace, the arm was placed on a retractable, movable pole hanging from the ceiling for the second scenario~(see \figref{fig:simulatedenv2}).
With this setup, the arm is able to approach most of the potential poses in the simulated room.
The simulated environments are available on GitHub\footnote{\url{https://github.com/Eruvae/ur_with_cam_gazebo}}.

We used simulated capsicum plants in the scenarios and determined the bounding boxes of the fruits in the local plant coordinate system to evaluate the results.
Additionally, the mesh of the fruits was converted into an octree to be able to directly compare the ROI nodes.
For the evaluation, we used the following metrics: \begin{itemize} 
	\item \textit{Number of detected ROIs}: Number of found clusters 
that can be matched with a ground truth cluster, which means that their center distance is smaller than~$20\,cm$.
	\item \textit{Cluster center distance}: Average distance of the detected cluster centers from the ground truth centers.
	\item \textit{Volume accuracy}: Average accuracy of the cluster volumes. 
The sizes of the axis-aligned 3D bounding boxes of the ROIs are compared, i.e., we determined the difference of the ground truth and the detected volume divided by the ground truth volume.
	\item \textit{Covered ROI volume}: Percentage of the total volume of the ground truth that was detected, considering the 3D bounding boxes. 
\end{itemize} The planner was given a total planning time of three minutes.

In order to show that the ROI targeted sampling is beneficial, we evaluated and compared two approaches: Our approach, which automatically switches between ROI targeted and exploration sampling, and pure exploration sampling at frontiers of occupied cells without considering ROIs, similar to \mbox{Monica\etal\cite{monica2018contour}}.
We tested the two utilities for viewpoint evaluation as described in \Cref{sec:approach_evaluation} for both approaches.
The proximity count utility is shortened with $U_P$, the unobserved voxel utility with $U_U$.
For each scenario, we executed 20 trials.
In order to show the statistical significance of the results, we performed a one-sided Mann-Whitney U test on the acquired samples.

\begin{figure} 	\centering
	\begin{subfigure}[b]{0.49\linewidth} 		\centering 		\includegraphics[width=\linewidth]{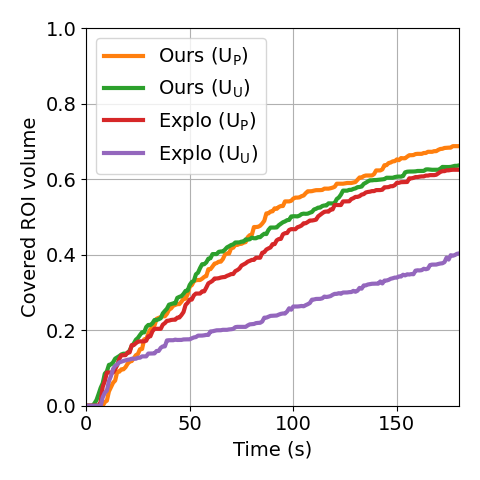} 		\caption{Covered ROI volume} 
		\label{fig:res_w14_covvol}
	\end{subfigure} 
	\begin{subfigure}[b]{0.49\linewidth} 		\centering 		\includegraphics[width=\linewidth]{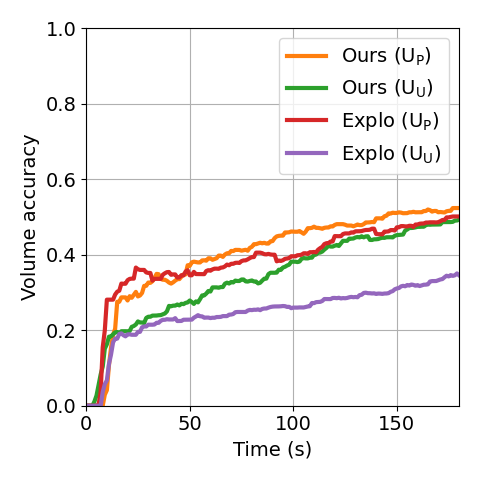} 		\caption{Volume accuracy} 
		\label{fig:res_w14_volacc}
	\end{subfigure} 	\hfill 	
	\caption{Results for Scenario 1~(\figref{fig:simulatedenv1}).
For each tested approach, 20 trials with a duration of three minutes each were performed.
The plots show the average results.
Our approach with $U_P$ performs the best, but the advantage over $U_U$ or exploration sampling with $U_P$ is minor. Exploration sampling with $U_U$ performs the worst.}
	\label{fig:res_w14}
\end{figure}

\begin{figure} 	\centering
	\begin{subfigure}[b]{0.49\linewidth} 		\centering 		\includegraphics[width=\linewidth]{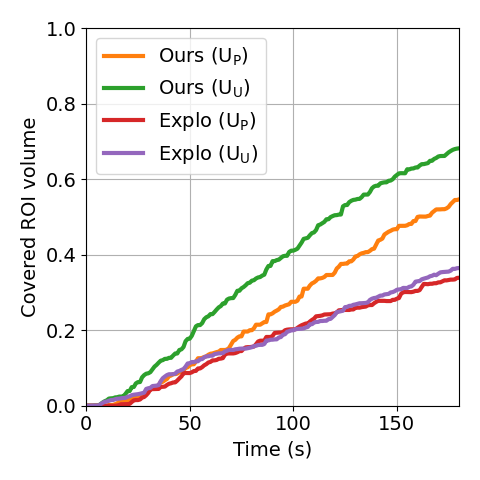} 		\caption{Covered ROI volume} 
		\label{fig:res_w19_covvol}
	\end{subfigure}
	\begin{subfigure}[b]{0.49\linewidth} 		\centering 		\includegraphics[width=\linewidth]{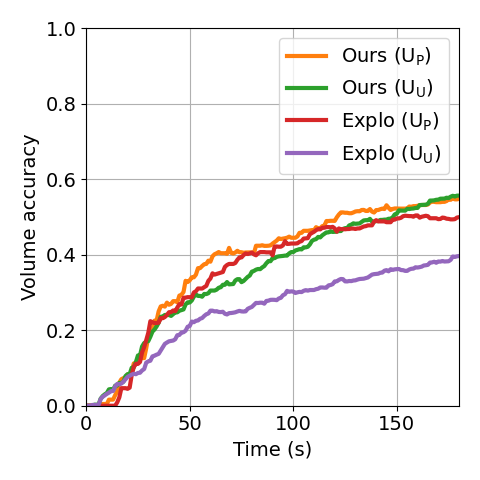} 		\caption{Volume accuracy} 
		\label{fig:res_w19_volacc}
	\end{subfigure} 	\hfill 	 
	\caption{Results for Scenario 2~(\figref{fig:simulatedenv2}).
Like in Scenario 1, 20 trials were performed for each approach and the plots show the average results. Here, our approach with $U_U$ has the best results, followed by the approach with $U_P$ utility. Exploration sampling with both utilities performs significantly worse.} 
	\label{fig:res_w19}
\end{figure}

\setlength{\extrarowheight}{2pt} 

\begin{table*} 	\centering 	\begin{tabularx}{\linewidth}{| c | X | X | X | X | X |} \cline{3-6} 		\multicolumn{2}{c|}{} & Ours-$U_P$ & Ours-$U_U$ & Explo-$U_P$ & Explo-$U_U$ \\ \cline{3-6}\hline 		\parbox[t]{2mm}{\multirow{4}{*}{\rotatebox[origin=c]{90}{Scen.
1}}} 		& \# Detected ROIs & 13.2 $\pm$ 0.7 & 13.8 $\pm$ 0.4 & 12.8 $\pm$ 1.1 & 13.4 $\pm$ 1.4 \\ \cline{2-6} 		& Covered ROI volume & \textbf{0.69 $\pm$ 0.07} & \textbf{0.64 $\pm$ 0.12} & \textbf{0.63 $\pm$ 0.11} & 0.40 $\pm$ 0.12 \\ \cline{2-6} 
		
		& Center distance (cm) & 2.41 $\pm$ 0.48 & 2.44 $\pm$ 0.45 & 2.23 $\pm$ 0.54 & 2.80 $\pm$ 0.47 \\ \cline{2-6} 		& Volume accuracy & \textbf{0.52 $\pm$ 0.07} & \textbf{0.49 $\pm$ 0.10} & \textbf{0.50 $\pm$ 0.21} & 0.35 $\pm$ 0.08 \\ \hline \hline 		\parbox[t]{2mm}{\multirow{4}{*}{\rotatebox[origin=c]{90}{Scen.
2}}} 		& \# Detected ROIs & 23.3 $\pm$ 2.9 & \textbf{27.0 $\pm$ 2.4} & 17.4 $\pm$ 3.9 & 24.8 $\pm$ 4.1 \\ \cline{2-6} 		& Covered ROI volume & 0.55 $\pm$ 0.11 & \textbf{0.68 $\pm$ 0.10} & 0.34 $\pm$ 0.08 & 0.36 $\pm$ 0.13 \\ \cline{2-6} 		& Center distance (cm) & 2.31 $\pm$ 0.45 & 2.21 $\pm$ 0.35 & 2.75 $\pm$ 0.51 & 2.96 $\pm$ 0.43 \\ \cline{2-6} 		& Volume accuracy &\textbf{0.55 $\pm$ 0.07} & \textbf{0.56 $\pm$ 0.08} & \textbf{0.50 $\pm$ 0.09} & 0.40 $\pm$ 0.08 \\ \hline 	\end{tabularx} 	\caption{Quantitative results over 20 trials.
Bold values show a significant improvement compared to the other approaches.
} 
	\label{tab:res_table}
\end{table*} 

\tabref{tab:res_table} shows the quantitative results for both scenarios.
\figref{fig:res_w14} illustrates the averaged ROI volume and volume accuracy over the three minutes planning time in Scenario~1.
As can be seen, our approach performs the best with $U_P$.
It has the most covered volume and highest volume accuracy, although the values are only slightly better than for our approach with $U_U$ or exploration sampling with the $U_P$.
An explanation for that could be that the ROI targeted sampling already ensures viewpoints nearby and directed towards ROIs, so using the proximity utility does not bring any additional advantage.
For exploration sampling, on the other hand, $U_P$ can be used to prefer viewpoints near ROIs, since this is not considered during sampling.

As can be seen from the results in the second scenario~(\figref{fig:res_w19}), our approach works best with $U_U$.
A reason why the simpler metric achieves better results could be the computation time.
The distance to the nearest ROI has to be computed in each planning step to enable determining the weight for $U_P$.
Thus, in the larger environment, computational costs are higher.
Using the simpler metric allows a faster evaluation of the viewpoints.
Therefore, more poses can potentially be reached within the given planning time.
\tabref{tab:res_table} shows that our approach with $U_U$ achieves a significantly higher number of detected ROIs and a significantly higher covered ROI volume than pure exploration sampling.

These experiments demonstrate that our sampling approach outperforms exploration sampling, which suggests that considering detected ROIs during planning improves the efficiency of gaining information about ROIs.
Furthermore, it enables using a simple metric for viewpoint evaluation, which can have computational advantages, especially in large environments.

However, while most fruits were detected successfully, the accuracy of the determined volume is limited, with average values of 0.52 and 0.56 for the best approach in the trials.
One reason is the naive clustering approach in combination with the relatively low resolution of the octree of 1 cm.
However, a higher resolution is not feasible for online planning, and the gathered information is sufficient to plan viewpoints.
In the future, we plan to combine our planning approach with a method to generate high-resolution 3D models from the recorded point clouds.
The ROI information will then be used to find the location of fruits, whereas the high-resolution model will be used to determine a more accurate volume.

\subsection{Real-World Glasshouse Experiment}

\begin{figure}
	\centering
	\includegraphics[width=\linewidth]{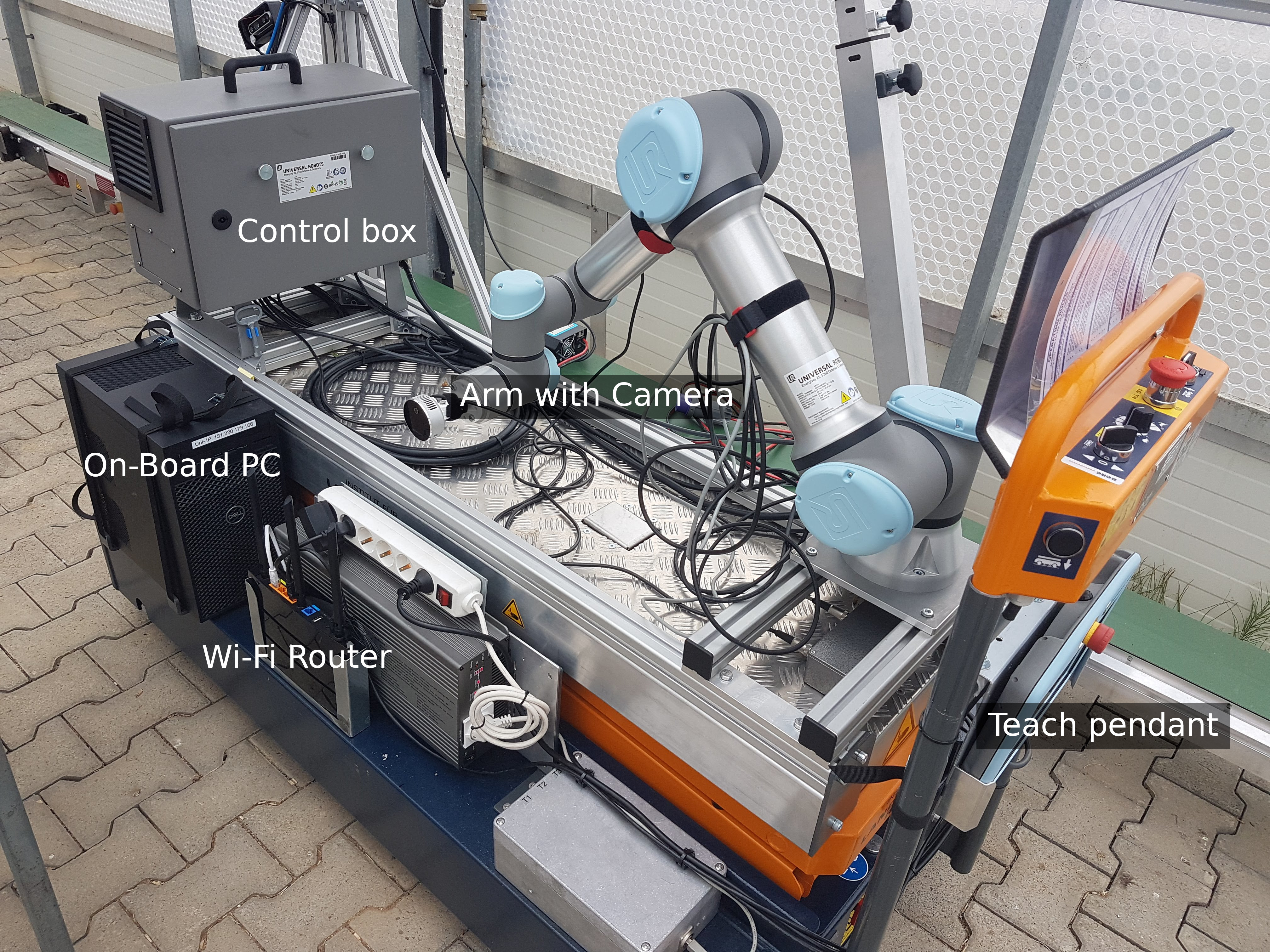}
	\caption{Pipe-rail trolley with a robotic arm equipped with a RealSense L515 Lidar sensor.}
	\label{fig:arm_rw_photos}
\end{figure}

For real-world experiments, we deployed the robotic arm to plan viewpoints in a capsicum glasshouse.
The arm is equipped with a RealSense L515 Lidar sensor, which is used as RGB-D input for the planner.
The arm was hereby placed on top of a pipe-rail trolley~(see \figref{fig:arm_rw_photos}), which has a pneumatically actuated scissor-lift to lift the platform up to $3\,m$.
The platform height and movement along the pipes can be controlled.
This enables us to map a complete row autonomously utilizing the planning framework, which we plan to do in the future.
For now, the planner was set to only control the arm in the experiments.
It was given some time to explore the environment at the current position of the trolley, then the planner was paused and the trolley moved forward manually by the length of the platform using the control GUI.
Since the trolley publishes its position estimated from a wheel encoder, the planner can combine a whole row of plants in a single map.

\begin{figure}
	\centering
	\begin{subfigure}[b]{0.49\linewidth} 		\centering 		\includegraphics[width=\linewidth]{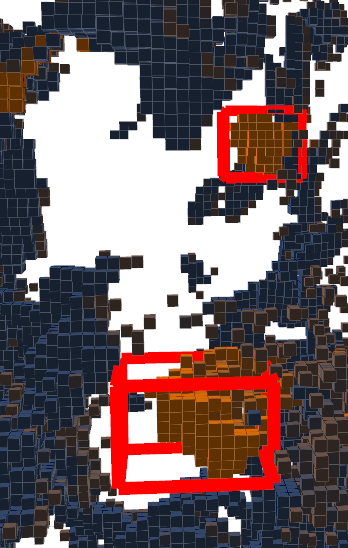} 		\caption{ROI octree} 
		\label{fig:glasshouse_rois}
	\end{subfigure}
	\begin{subfigure}[b]{0.49\linewidth} 		\centering 		\includegraphics[width=\linewidth]{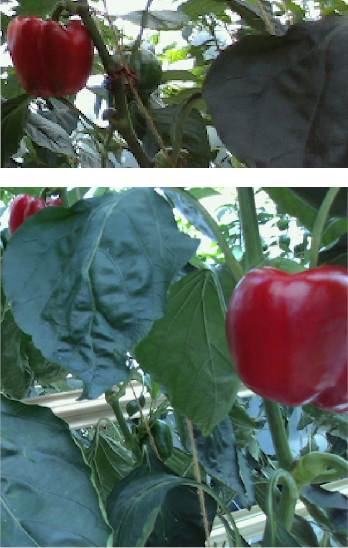} 		\caption{Captured photos}
		\label{fig:glasshouse_photo}
	\end{subfigure}
	\caption{Segment of the recorded row in the glasshouse.
		Left: Visualization of the ROI octree during planning.
		Orange cells are ROIs, blue cells
		indicate other occupied regions.
		The red bounding boxes mark the detected fruits.
		Right: Pictures of these fruits captured during planning. One of the fruits is mostly occluded by a leaf in the bottom picture, but visible in full in the top picture.} 
	\label{fig:glasshouse_results}
\end{figure}

\Cref{fig:glasshouse_results} shows the resulting map for part of a row in one of the trials.
Since there is no ground truth available and the number of performed trials was limited, no numerical analysis could be performed.
However, the experiment still shows that the planning approach is viable for real-world challenges and fruit regions can be estimated.
The shown segment shows two red peppers that were successfully recognized and clustered during planning.
Due to false detections of the neural network, the full octree also contains some wrongly marked fruits, and the map is less accurate than in the simulations due to noise.
Nevertheless, our approach was able to sample reasonable viewpoints to move the arm.

\section{Conclusions}
\label{sec:concl}

We introduced a novel viewpoint planning approach that detects ROIs in the environment and samples viewpoint candidates from them in order to achieve a high ROI coverage.
Our planning framework allows switching between different planning modes, depending on the stage of coverage, i.e., our system automatically switches between ROI targeted and general exploration sampling.
We demonstrated in simulated experiments that our sampling approach outperforms a method that does not consider the information about ROIs with respect to the number of correctly detected ROIs and the covered ROI volume.
The ROI targeted viewpoint sampling enables good results even with a simple metric for viewpoint evaluation, which can be an advantage in larger maps, where more complex metrics become computationally expensive.
We also showed that our planner can be used in a real-world environment by demonstrating its use on a robotic platform in a commercial glasshouse environment with capsicum plants.

\bibliographystyle{IEEEtran}
\bibliography{refs}

\end{document}